\pdfoutput=1

\documentclass[11pt]{article}

\usepackage[final]{acl}

\usepackage{times}
\usepackage{latexsym}

\usepackage[T1]{fontenc}
\usepackage[utf8]{inputenc}
\usepackage{microtype}
\usepackage{inconsolata}
\usepackage{graphicx}

\title{Infusing Knowledge into Large Language Models with Contextual Prompts}

\author{Kinshuk Vasisht\\
  University of Delhi \\
  New Delhi, India \\
  \texttt{kinshuk.mcs21@cs.du.ac.in} \\\And
  Balaji Ganesan \\
  IBM Research India\\
  Bengaluru, India \\
  \texttt{bganesa1@in.ibm.com} \\\AND
  Vikas Kumar \\
  University of Delhi\\
  New Delhi, India \\
  \texttt{vikas@cs.du.ac.in} \\\And
  Vasudha Bhatnagar \\
  University of Delhi \\
  New Delhi, India \\
  \texttt{vbhatnagar@cs.du.ac.in} \\
}

\begin{document}
\maketitle
\begin{abstract}
Knowledge infusion is a promising method for enhancing Large Language Models for domain-specific NLP tasks rather than pre-training models over large data from scratch. These augmented LLMs typically depend on additional pre-training or knowledge prompts from an existing knowledge graph, which is impractical in many applications. In contrast, knowledge infusion directly from relevant documents is more generalisable and alleviates the need for structured knowledge graphs while also being useful for entities that are usually not found in any knowledge graph. With this motivation, we propose a simple yet generalisable approach for knowledge infusion by generating prompts from the context in the input text. Our experiments show the effectiveness of our approach which we evaluate by probing the fine-tuned LLMs.
\end{abstract}

\section{Introduction}

Unifying Large Language Models (LLMs) and Knowledge Graphs (KGs) is an active area of research for several reasons. \citet{pan2023unifying} presents a survey of different approaches, including knowledge infusion into Large Language Models from knowledge bases. Common techniques for infusing knowledge involve pre-training over a factually-rich corpus prepared from structured knowledge bases or learning soft knowledge prompts pertaining to entities using factual triples from a knowledge base for improving entity-specific inference \cite{santos2022knowledge}.

Infusing knowledge directly into the model from knowledge bases, though more efficient than re-training LLMs from scratch, is unrealistic in many real-world applications. Maintaining knowledge graphs about customers, organizations and events mentioned in documents is cumbersome and incurs an overhead of maintaining privacy. Further, all entities are not equal in knowledge bases, with some entities being highly under-represented. This degrades the performance over downstream tasks for such entities, post-knowledge infusion, due to lack of sufficient knowledge. Such entities may be more frequent in domain-specific corpora, which can provide the relevant context for knowledge infusion. As an example, while entities such as \textit{Barack Obama} and \textit{Michelle Obama} are well known and represented in structured knowledge bases such as Wikidata, entities such as \textit{Michelle Obama}'s great aunt \textit{Robbie Shields Terry}, being relatively lesser known, exist only in the Wikipedia page of the subject and not as an entity in Wikidata.

\begin{figure}
  \includegraphics[width=\columnwidth]{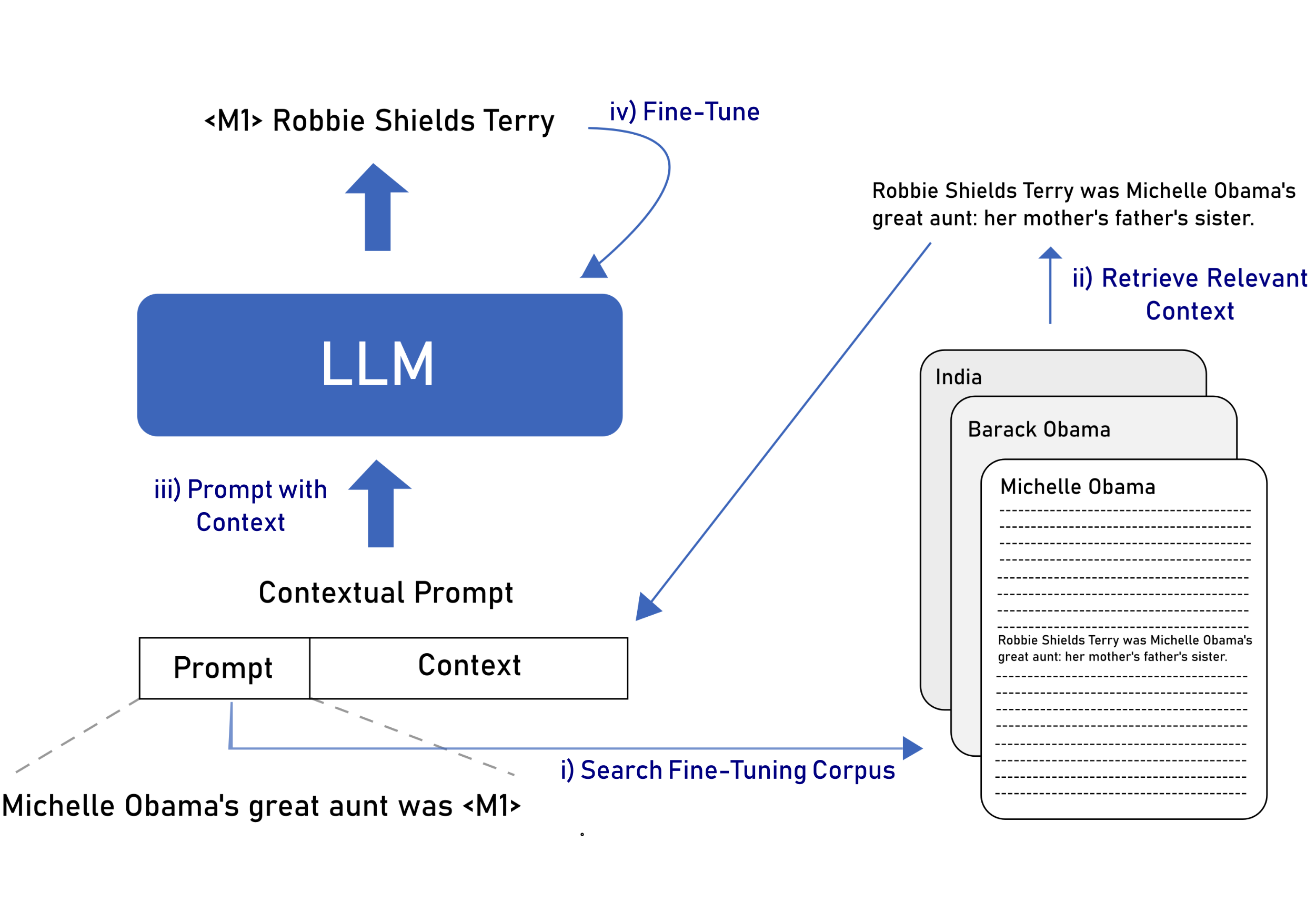}
  \caption{Contextual prompts to infuse knowledge about entities into Large Language Models}
  \label{fig:contextual_prompts}
\end{figure}

Motivated by the above, we propose to exploit contextual text from a relevant domain-specific corpus to infuse knowledge into Large Language Models. Infusing knowledge directly from documents without having to create knowledge graphs is not only efficient, but also more general. Figure \ref{fig:contextual_prompts} outlines the four steps of our proposed approach. Given an input prompt composed of a task instance and accompanying instructions, we retrieve relevant context from an indexed corpus by identifying sentences mentioning involved entities. Knowledge is then infused into a pre-trained model by fine-tuning over input prompts augmented with the identified context. The knowledge-infused model obtained after fine-tuning can be leveraged for knowledge-intensive downstream tasks. We compare and contrast our knowledge-infusion method against other infusion techniques involving prompting with factual triples \cite{moiseev-etal-2022} and natural sentences from triples \cite{agarwal-etal-2021-knowledge}.

Our approach offers significant advantages over existing knowledge infusion methods. The proposal alleviates the need for structured knowledge sources by relying solely on domain corpora. Using fine-tuning makes our approach simple, scalable, and employable in low-resource settings where excessive compute resources and data sources are unavailable. The method permits seamless integration of structured knowledge graphs, if available, to further enhance the identification of relevant context from corpora via entity linking and disambiguation. Our method is also extensible to other modalities like tabular data and graphs.

\section{Related Works}

\citet{petroni2019language} introduced the idea of language models being knowledge bases. Since then there have been continuous efforts to both extract facts from LLMs and also to infuse language models with facts from knowledge bases.

\citet{agarwal-etal-2021-knowledge} generated natural language sentences from triples and additionally pre-trained large language models with the generated sentences. \citet{moiseev-etal-2022} directly infuse triples into LLMs without generating sentences from the triples. \citet{agarwal-etal-2022} infuse triples from domain-specific knowledge graphs into T5 models. These approaches show the interchangeable nature of knowledge infused from both triples and sentences containing the triples.

\citet{santos2022knowledge} proposed \textit{Knowledge Prompts}, where prompts are learnt for the most frequently occurring entities in Wikidata. For every triple, a prompt is initialised with random values and then updated via gradient descent on the triples mask prediction task. Other methods ~\cite{wang2021k,diao2023mixtureofdomainadapters} explore the use of adapters for knowledge infusion with parameter efficient fine-tuning. ~\citet{de2021editing} \& \citet{zhong2023mquake} discuss related ideas in knowledge editing.

\emph{Existing approaches assume the existence of a well-populated KG, and hence suffer from limitations concerning practicality in a real-world setting. For example, a new customer entity, a new product or new terms in a news article or a court judgment may not exist in the KG.}

\section{Knowledge Infusion with context}

\begin{table*}[!ht]
    \centering
    \resizebox{\textwidth}{!}{
    \begin{tabular}{llrrrrr}
    \hline

    \textbf{Dataset} & \textbf{Model} & \textbf{Hits@1} $\uparrow$ & \textbf{Hits@5} $\uparrow$ & \textbf{Hits@10} $\uparrow$ & \textbf{AED} $\downarrow$ & \textbf{MRR} $\uparrow$ \\
    \hline
    KELM-TEKGEN & google/flan-t5-small & 0.019 & 0.036 & 0.045 & 18.75 & 0.024\\ 
        & google/flan-t5-base & 0.047 & 0.063 & 0.095 & 85.5 & 0.055 \\ 
        & google/flan-t5-large & 0.082 & 0.102 & 0.138 & 109.5 & 0.088  \\ 
        & flan-t5-small-fine-tuned & 0.528 & 0.535 & 0.541 & 96.75 & 0.538 \\ 
        & flan-t5-base-fine-tuned & 0.514 & 0.520 & 0.539 & 83.25 & 0.525 \\
        & flan-t5-small-fine-tuned-w-context & 0.800 & 0.801 & 0.804 & 2.75 & 0.805 \\
        & flan-t5-base-fine-tuned-w-context & \textbf{0.825} & \textbf{0.825} & \textbf{0.833} & \textbf{0.75} & \textbf{0.827} \\
    \hline
    TACRED & google/flan-t5-small & 0.004 & 0.006 & 0.006 & 84.75 & 0.005 \\ 
        & google/flan-t5-base & 0.004 & 0.014 & 0.018 & 9.75 & 0.008 \\ 
        & google/flan-t5-large & 0.034 & 0.044 & 0.060 & 22.50 & 0.039 \\ 
        & flan-t5-small-fine-tuned & 0.366 & 0.368 & 0.39 & 50.25 & 0.376 \\ 
        & flan-t5-small-fine-tuned-w-context & 0.782 & 0.782 & 0.784 & \textbf{3.75} & 0.788 \\
        & flan-t5-base-fine-tuned-w-context & \textbf{0.818} & \textbf{0.820} & \textbf{0.824} & 5.25 & \textbf{0.823} \\
    \hline
    Re-TACRED & google/flan-t5-small & 0.000 & 0.010 & 0.016 & 66.00 & 0.005 \\ 
        & google/flan-t5-base & 0.006 & 0.016 & 0.028 & 28.50 & 0.010 \\ 
        & google/flan-t5-large & 0.052 & 0.070 & 0.084 & 5.25 & 0.060 \\ 
        & flan-t5-small-fine-tuned & 0.352 & 0.366 & 0.406 & 15.75 & 0.370 \\ 
        & flan-t5-small-fine-tuned-w-context & 0.798 & 0.798 & 0.800 & 6.00 & 0.805 \\ 
        & flan-t5-base-fine-tuned-w-context & \textbf{0.846} & \textbf{0.846} & \textbf{0.850} & \textbf{0.00} & \textbf{0.852} \\
    \hline
    \end{tabular}
    }
    \caption{Flan-T5 performance on relation prediction task on KELM-TEKGEN, TACRED and Re-TACRED datasets.}
    \label{tab:expr_triple_prediction}
\end{table*}

Our motivation is to create a knowledge-prompting approach that draws on documents and large language models rather than only KGs. Accordingly, our approach leverages relevant context retrieved from a domain-specific fine-tuning corpus to infuse knowledge into a pre-trained language model. This is in contrast to approaches utilizing factual triples \cite{moiseev-etal-2022} or corresponding natural sentences \cite{agarwal-etal-2021-knowledge} from knowledge bases, or soft knowledge prompts prepared in an entity-specific manner \cite{santos2022knowledge} as the source for knowledge to be infused.

We prepare a pre-trained model by infusing knowledge using full fine-tuning over a specific downstream task, such as tail prediction in triples, question answering, or translation. For this purpose, along with a suitable dataset comprising of downstream task instances, we also identify a domain-specific fine-tuning corpus composed of relevant documents that contain information pertaining to involved entities, which serves as our source of knowledge. Primarily, we formulate instances from a training dataset into prompts, by prepending brief task-specific instructions alongside the instance data. For example, for a tail-prediction task, alongside a factual triple without the tail entity, an instruction describing the task of tail-prediction is prepended. Next, for a given prompt, we identify named entities present in the prompt and retrieve relevant context from the domain-specific fine-tuning corpus. This context is composed of useful information that could provide knowledge about the involved entities to aid in enhancing performance for the task. We infuse knowledge from the context surrounding the entity by augmenting the context alongside the text prompt. For each entity, our contextual information is the phrase, sentence, or paragraph surrounding the entity in a document index. Since we are doing full fine-tuning, the model parameters are updated, and the error is propagated based on the task.

During inference, task-specific prompts comprising of task instructions and the query to process are given, for which responses are generated based on the knowledge infused. Unlike the fine-tuning stage, relevant context is not retrieved and utilized for inference.

As an example, for the tail and link prediction task, a dataset for fine-tuning comprising triples is identified, alongside which we identify a relevant domain-specific corpus for the source of knowledge. We formulate the factual triple as a prompt for the LLM, post-masking one of the entities or the relation at random. For the unmasked entities and relation, we retrieve relevant context from the corpus and append it to the generated input prompt to obtain the contextual prompt finally used to fine-tune the LLM. Post fine-tuning, at inference the model is prompted with factual triples formulated as a prompt in the same format but without retrieved context.

Unlike the approach in \citet{santos2022knowledge}, neither entity linking via an external knowledge graph nor inference based on information/prompts derived from structured knowledge bases is performed. Further, the prompt formulation contrasts from \citet{agarwal-etal-2022} and \citet{saxena-etal-2022-sequence}, where for entity and link prediction explicit mentions of the masked positions is present and additional context is not leveraged during fine-tuning.

In this work, we directly utilize retrieved-context alongside the input prompt, which is comprised of tokens and is therefore discrete in nature as opposed to soft prompts in a continuous latent space. To include larger amounts of information and bypass the context length restrictions of LLMs, contextual prompts can be prepared based on context embedded as soft prompts, allowing for further improvements to efficiency and performance. Further, for tasks where structured knowledge sources or information in other modalities, such as tabular data is available, our approach could be extended to leverage the information from these sources and create appropriate contextual prompts for knowledge infusion.

\section{Experiments}

We conduct experiments with pre-trained Flan-T5 models \cite{chung-etal-2022} to demonstrate the role of contextual text as discrete text prompts. We fine-tune the model on two downstream tasks, tail prediction in triples and question answering. Tail prediction, as the name indicates, predicts the object in subject, predicate, and object triples.

Our experiments leverage the following variants of Flan-T5 models: Flan-T5-Small with 80M parameters and Flan-T5-base with 250M parameters. Fine-tuning and inference were performed in FP-16 mixed-precision mode \cite{DBLP:journals/corr/abs-1710-03740} over the task-specific datasets using a single Nvidia A4000 GPU with 16GBs of VRAM available. Additionally, gradient accumulation and gradient check-pointing \cite{DBLP:journals/corr/ChenXZG16} techniques were employed to reduce memory consumption.

\subsection*{Relation Prediction}

We perform the relation extraction experiment on four datasets. KELM-TEKGEN dataset from \citet{agarwal-etal-2021-knowledge} has over 15 million Wikidata triples and sentences generated with those triples, from which we leverage a sample of 1 million triples partitioned in a ratio of 9:1 for fine-tuning and evaluation. TACRED \citet{zhang-etal-2017} is a well-known relation extraction dataset comprising of about 60,000 factual triples with associated text. Our experiments leverage the subset of 18,000 triples for which relations are known paritioned as before for fine-tuning and evaluation. Re-TACRED is a variant of TACRED introduced by \citet{stoica-etal-2021} with a larger number of triples with known relations. We compare the results using metrics commonly utilized for entity and link prediction in Knowledge Graph literature, notably Mean Reciprocal Rank (MRR) and Hits@K (\citet{voorhees-tice-2000-trec}, \citet{Ali_2022}) with varying values of K: 1,5 and 10. We also use an approximation of the Graph Edit Distance metric as in \citet{swamy-etal-2021} for comparing the generated knowledge graphs against the ground truth, abbreviated as AED. 

Table~\ref{tab:expr_triple_prediction} summarizes the results of our experiments. We observe that all the fine-tuning methods perform better the base Flan-T5 models. Although we observe this across different sizes of Flan-T5 models we compare against (small, base and large), we believe the knowledge available to the model is more relevant in these experiments rather than the size of the models. This is discussed in \citet{agarwal-etal-2022}, which investigates the role of model size on the model's capacity to retain knowledge.

Among the fine-tuning methods, our introduction of contextual text as discrete text prompts significantly improves the performance across all datasets and metrics.

\subsection*{Soft Prompts}

While the use of context significantly improves performance, to justify the efficacy of the method towards knowledge infusion, notably building of knowledge regarding entities present in the text and/or triples, we conduct additional investigatory experiments evaluating entity knowledge developed by the model as a consequence of the infusion process. We learn differentiable `contextual' prompts termed as soft prompts, specific to each entity of the training corpus to aid inference in downstream tasks. We compare the performance of adapter models utilizing these contextual prompts over tail prediction and relation extraction against full fine-tuning with and without context in Table \ref{tab:expr_kelm_matched}.

\begin{table}[!ht]
    \resizebox{\columnwidth}{!}{
    \begin{tabular}{lrrrrr}
        \hline
        \textbf{Model} & \textbf{Hits@1} & \textbf{AED} & \textbf{MRR} \\ \hline
        google/flan-t5-base & 0.050 & 11.250 & 0.056 \\ 
        flan-t5-base-fine-tuned & 0.441 & 17.250 & 0.452 \\ 
        flan-t5-base-fine-tuned-w-context & 0.710 & 10.500 & 0.717 \\ 
        flan-t5-base-fine-tuned-soft-prompt & 0.176 & 6.750 & 0.174 \\ \hline
    \end{tabular}
    }
    \caption{Flan-T5 performance on relation prediction on a subset of KELM-TEKGEN test data with entities known from the training phase}
    \label{tab:expr_kelm_matched}
\end{table}

\subsection*{Question Answering Task Evaluation}

To evaluate our approach on tasks other than relation extraction, we use a closed-book open-domain question-answering task on the Trivia QA dataset \citet{2017arXivtriviaqa}. We use the rc-wikipedia partition of the dataset and evaluate the models fine-tuned over triples from the KELM corpus. The models are fine-tuned again for the question-answering task for a maximum of 10K steps and a batch size of 128.

For each question, we identify entities and perform a lookup in the KELM corpus to retrieve aligned sentences for the entity in question. The retrieved context is then used alongside for prompting. We evaluate the previously selected models after fine-tuning them over the question-answering task as described above using the same settings as described in the previous experiment. Table \ref{tab:expr_trivia_qa_w_context} summarizes the exact match scores observed.

\begin{table}[!ht]
    \centering
    \resizebox{\columnwidth}{!}{
    \begin{tabular}{lr}
        \hline
        \textbf{Model} & \textbf{Exact Match (\%)} \\ \hline
        google/flan-t5-small & 27.1 \\
        flan-t5-small-fine-tuned & 16.7 \\
        flan-t5-small-fine-tuned-w-context & 20.0 \\ \hline
    \end{tabular}
    }
    \caption{Flan-T5 performance on question-answering over the TriviaQA dataset}
    \label{tab:expr_trivia_qa_w_context}
\end{table}

\subsection*{Impact of Variation in Context Size}

In order to investigate the role of the context towards knowledge infusion and prediction, we conduct experiments to examine the impact of change in context length (in terms of either a change in the number of sentences or the length in terms of the number of tokens). This provides an insight into whether the models are able to produce satisfactory results only in the presence of a context that directly describes the entity to be predicted or whether the predictions are a result of the understanding and insight provided by the context. Results as summarized in table \ref{tab:expr_retacred_var_len} demonstrate improvements with the use of larger contexts with more information.

\begin{table}[!ht]
    \centering
    \resizebox{\columnwidth}{!}{
    \begin{tabular}{lrrrrr}
    \hline
    \textbf{Model} & \textbf{Hits@1} & \textbf{AED} & \textbf{MRR} \\ \hline
flan-t5-small & 0.798 & 6.00 & 0.805 \\ 
flan-t5-small-longer-context & 0.836  & 2.25 & 0.842 \\
flan-t5-base & 0.846 & 0.00 & 0.852 \\
flan-t5-base-longer-context & 0.914 & 3.75 & 0.917 \\\hline
    \end{tabular}
    }
    \caption{Impact of context lengths over entity and relation prediction performance}
    \label{tab:expr_retacred_var_len}
\end{table}

\subsection*{Legal Corpus}

While the experiments in Table \ref{tab:expr_triple_prediction} are predominantly on news or web knowledge, we conducted a similar experiment on a legal knowledge graph and related documents from \citet{dhani-etal-2021}. This is one of the example applications where we expect the domain knowledge graphs to be non-existent or incomplete. As shown in Table \ref{tab:expr_legal}, our contextual prompts seem to work in this scenario, though the improvement is modest.

\begin{table}[!ht]
    \centering
    \resizebox{\columnwidth}{!}{
    \begin{tabular}{lrrrr}
    \hline
    \textbf{Model} & \textbf{Hits@10} & \textbf{AED} & \textbf{MRR} \\ \hline
        flan-t5-base & 0.245 & 47.25 & 0.200 \\
        flan-t5-base-w-context & 0.290 & 45.00 & 0.217 \\ \hline
    \end{tabular}
    }
    \caption{Flan-T5 performance on entity and link prediction over the Legal KG, with corresponding judgments used as the context source}
    \label{tab:expr_legal}
\end{table}

\section*{Limitations}
In this work, we have limited our experiments to full fine tuning and examining the role of contextual text in helping LLMs surface the knowledge about entities. We have not compared our method with other knowledge infusion methods like additional pre-training, adapters and soft prompts. Further, an evaluation of knowledge propagation, such as in \citet{onoe-etal-2023} has not been performed, which could be an interesting direction to explore in future research for understanding the extent of knowledge infused in models. Our work is more relevant to applications where an external knowledge graph is unavailable or is not well populated. We show that contextual text can be used in lieu of knowledge graphs in such applications.

\section*{Conclusion}
We propose an alternative method to infuse knowledge into Large Language Models (LLMs) that does not assume the existence of a Knowledge Graph (KG). We use a search index to provide relevant sentences to be used as context alongside input prompts during fine-tuning for knowledge infusion. Results over relation extraction and tail predictions tasks demonstrate improved extents of knowledge infusion with the use of context.

\section*{Acknowledgements}
This work was done by the first author as part of his graduate coursework, being a part of the Legal Text Analytics group at the University of Delhi. We thank the Legal Text Analytics group and the University of Delhi for providing this opportunity. The second author participated in this work as part of the Global Remote Mentorship program of IBM. We thank IBM University Relations for their support. We thank the reviewers for their valuable suggestions.

\bibliography{references}

\end{document}